\documentclass{article} % For LaTeX2e
\usepackage{aistats2020}

\usepackage{amsmath,amsfonts,bm}

\def\eqref#1{equation~\ref{#1}}
\def\1{\bm{1}}

\DeclareMathAlphabet{\mathsfit}{\encodingdefault}{\sfdefault}{m}{sl}
\SetMathAlphabet{\mathsfit}{bold}{\encodingdefault}{\sfdefault}{bx}{n}

\usepackage[utf8]{inputenc} % allow utf-8 input
\usepackage[T1]{fontenc}    % use 8-bit T1 fonts
\usepackage{hyperref}       % hyperlinks
\usepackage{url}            % simple URL typesetting
\usepackage{booktabs}       % professional-quality tables
\usepackage{amsfonts}       % blackboard math symbols
\usepackage{nicefrac}       % compact symbols for 1/2, etc.
\usepackage{microtype}      % microtypography

\usepackage[utf8]{inputenc} % allow utf-8 input
\usepackage[T1]{fontenc}    % use 8-bit T1 fonts
\usepackage{url}            % simple URL typesetting
\usepackage{booktabs}       % professional-quality tables
\usepackage{amsfonts}       % blackboard math symbols
\usepackage{nicefrac}       % compact symbols for 1/2, etc.
\usepackage{microtype}      % microtypography

\usepackage{natbib}

\usepackage{tikz}
\usetikzlibrary{fit,positioning}

\usepackage{filecontents}
\usepackage{amsmath}
\usepackage{array}
\usepackage{stfloats}

\usepackage{graphicx}
\usepackage{caption}
\usepackage{subcaption}

\usepackage{capt-of}

\usepackage{tikz}
\usetikzlibrary{fit,positioning}

\usepackage{multirow}
\usepackage{mathrsfs}

\usepackage{multirow}

\usepackage{enumitem}
\setlist{nolistsep}

\usepackage{adjustbox}

\usepackage{algorithm}
\usepackage{algorithmic}

\newcommand\ie {{\it i.e., }}

\usepackage{dsfont}

\begin{document}

\twocolumn[

\aistatstitle{Robust Ordinal VAE: Employing Noisy 
\\
Pairwise Comparisons for Disentanglement}

\aistatsauthor{ Junxiang Chen \And Kayhan Batmanghelich }

\aistatsaddress{ Department of Biomedical Informatics \\University of Pittsburgh \And  Department of Biomedical Informatics \\ University of Pittsburgh } ]

\begin{abstract}
%\input{sections/abstract_v1.tex}
% !TEX root = ../main.tex

Recent work by~\cite{locatello2018challenging} has shown that an inductive bias is required to disentangle factors of interest in Variational Autoencoder (VAE). Motivated by a real-world problem, we propose a setting where such bias is introduced by providing pairwise ordinal comparisons between instances, based on the desired factor to be disentangled. For example, a doctor compares pairs of patients based on the level of severity of their illnesses, and the desired factor is a quantitive level of the disease severity. In a real-world application, the pairwise comparisons are usually noisy. Our method, Robust Ordinal VAE (ROVAE), incorporates the noisy pairwise ordinal comparisons in the disentanglement task. We introduce non-negative random variables in ROVAE, such that it can automatically determine whether each pairwise ordinal comparison is trustworthy and ignore the noisy comparisons. Experimental results demonstrate that ROVAE outperforms existing methods and is more robust to noisy pairwise comparisons in both benchmark datasets and a real-world application.

\end{abstract}

% !TEX root = ../main.tex

\section{Introduction}
\label{sec:introduction}
Disentanglement is the task of recovering the latent explanatory factors of variations in the data \citep{bengio2013representation}.
In recent years, several unsupervised disentanglement learning methods have been proposed based on the variants of Variational Autoencoder (VAE) \citep{higgins2017beta, kim2018disentangling, chen2018isolating, lopez2018information}. However, \citet{locatello2018challenging} demonstrated with theory and experiments that unsupervised disentanglement learning is fundamentally impossible if no inductive biases are provided. Therefore, various forms of supervision have been proposed to improve the performance of disentanglement \citep{narayanaswamy2017learning,kulkarni2015deep, bouchacourt2018multi}.  In this paper,  we propose a disentanglement method, we call Robust Ordinal VAE (ROVAE), to solve the disentanglement problem, where supervision is provided based on noisy pairwise comparisons. 

Our method is motivated by a real-world scenario, where an uncertain user introduces supervision in the form of noisy pairwise comparison. For example, we are interested in disentangling the severity of a disease, but an existing measurement for disease severity is not available. A doctor can only provide insight by comparing pairs of patients, deciding whether one patient has a more severe disease than the other. Also, because of the limited expertise of the doctor, such comparison is subject to noise. The challenge is whether we can learn a latent representation that corresponds to the disease severity, making use of the data of the patients and the observed pairwise ordinal comparisons.

To solve this problem, we proposed a disentangling model based on the VAE framework, we call Robust Ordinal VAE(ROVAE), incorporating noisy pairwise ordinal comparisons. Since the provided pairwise ordinal labels are usually noisy, we introduce non-negative random variables to model the trustworthiness of each ordinal label. These random variables allow ROVAE to automatically determine whether a label is trustworthy, and to learn the latent representation based on more trustworthy labels while ignoring noisy labels. As shown in the empirical evaluations on several benchmark datasets and a real-world application, the proposed method is able to disentangle the factor of interest and it is more robust against noisy labels.

{\bf Contributions} \hspace{1mm}
The contributions of this paper are summarized as follows: (1)~We design a probabilistic model based on the VAE framework, incorporating the pairwise ordinal labels. 
(2)~We introduce trustworthiness scores to the model. This allows the model to automatically determine whether a pairwise comparison is trustworthy or not, and learn the latent representation based on more trustworthy labels while ignoring noisy labels.
(3)~We conduct extensive experiments on benchmark datasets and the real-world clinical dataset. Experimental results demonstrate that the proposed method is more robust against noisy pairwise ordinal labels.

% !TEX root = ../main.tex

\section{ Background }
\subsection{Unsupervised Disentanglement with Variational Auto-Encoder (VAE)}
\label{sec:background_VAE}
There have been several unsupervised disentangling methods based on Variational Auto-Encoder (VAE) \citep{higgins2017beta, kim2018disentangling, chen2018isolating, lopez2018information}. In general, the objective function of these methods can be summarized as follows:
\begin{equation}
    \label{equ:obj_VAE}
    \begin{aligned}
    \mathcal{L}_{VAE} = & 
    \min_{\boldsymbol{\theta}, \boldsymbol{\phi}}
    \hspace{2mm}
    \mathbb{E}_{ \mathbf{x}_n \sim p_{\text{data}} } 
        [\mathbb{E}_{q_{\boldsymbol{\phi}}(\mathbf{z}_n|\mathbf{x}_n) }[
        \mathcal{L}_{recon} (\widetilde{\mathbf{x}}_n,\mathbf{x}_n ;\boldsymbol{\theta}) ] 
    \\
        & + \mathcal{D}_{KL} 
        \left( q_{\boldsymbol{\phi}}(\mathbf{z}_n|\mathbf{x}_n) || p(\mathbf{z})
        \right) 
        + \beta \mathcal{R}( q_{\boldsymbol{\phi}}(\mathbf{z}_n|\mathbf{x}_n) 
        ],
    \end{aligned}    
\end{equation}
where $\mathbf{X} = \left\{ \mathbf{x}_n \right\}_{n =1}^{N}$ and $ \mathbf{Z} = \left\{ \mathbf{z}_n \right\}_{n =1}^{N}$ denote the observed samples and  the corresponding latent representations respectively, with $N$ being the number of samples. We use $\mathcal{L}_{recon} (\widetilde{\mathbf{x}}_n,\mathbf{x}_n ;\boldsymbol{\theta}) $ to represent the reconstruction loss, where $ \widetilde{ \mathbf{x} } $ represents the the reconstructed instance via decoding $\mathbf{z}_n$  and 
$ \boldsymbol{\theta}$  represents the parameter for the decoder network. We let $q_{\boldsymbol{\phi}}(\mathbf{z}_n|\mathbf{x}_n)$ be the encoder network that is parameterized by $\boldsymbol{\phi}$. We let $\mathcal{D}_{KL}(\cdot || \cdot)$ denote the KL divergence and $p(\mathbf{z})$ denote prior distribution for $\mathbf{z}$. In this paper, we let $p(\mathbf{z})$ be an isotropic unit Gaussian distribution. $ \mathcal{R}(\mathbf{Z}) $ is a regularization term that forces each dimension of $\mathbf{Z}$ to be independent with each other. $\beta$ is a non-negative hyper-parameter that controls the weight of $ \mathcal{R}(\mathbf{Z})$. Existing unsupervised disentanglement methods follow the same framework as introduced in Equation (\ref{equ:obj_VAE}), but each of them chooses a different $ \mathcal{R}(\mathbf{Z})$.

\subsection{Ordinal Losses}
\label{sec:related_work_ordinal}
Many existing methods make use of ordinal information with different motivations and formulations. In general, these methods can be divided into three categories \citep{chen2009ranking}: The pointwise approach \citep{cossock2008statistical, li2008mcrank}, which considers each single instance independently in the objective function; the pairwise approach \citep{herbrich1999support, freund2003efficient, burges2005learning}, which analyzes a pair of instances at a time in the loss function; and the listwise approach \citep{cao2007learning, xia2008listwise}, which considers entire list of instances simultaneously. 

In this paper, we focus on discussing pairwise ordinal loss. We let a set $\mathcal{J}$ represent the partially observed pairwise order between instances,  such that, $\mathcal{J} \subseteq \{ (i,j) | t_i>t_j\} $, where $\{t_n\}_{n = 1}^N$ represent the factor of interest. The existing methods represent the ordinal loss with hinge function~\citep{herbrich1999support}, exponential function~\citep{freund2003efficient} and logistic function~\citep{burges2005learning}, respectively. In this paper, we focus on discussing the logistic function case, such that
\begin{equation}
    \mathcal{L}_{ordinal} = - \sum_{(i, j) \in \mathcal{J}}
    \log \psi \left( \vphantom{\frac{1}{1}} f(\mathbf{x}_i) - f(\mathbf{x}_j) 
    \right) 
    \label{equ:logistic_y}
\end{equation}
where $\psi$ is the logistic function, \ie $\psi(t) = 1 / (1 + e^{-t})$, and $f(\mathbf{x})$ is a certain function that captures the ordinal information for instance $\mathbf{x}$. Note that Equation (\ref{equ:logistic_y}) treat all ordinal labels $(i, j) \in \mathcal{J}$ equally, and does not consider the case that some of the ordinal comparisons $t_i > t_j$ might be noisy.

% !TEX root = ../main.tex

\section{Methods}

In this section, we introduce our ROVAE model. We first introduce how we improve the robustness of the ordinal loss via introducing trustworthiness scores. Then we introduce how we incorporate the ordinal loss in the VAE framework. Later, we introduce how we approximate the posterior distributions of the random variables. Finally, we summarize the overall model, and analyze the mechanism how trustworthiness score help improve robustness of ROVAE.

\subsection{The Robust Ordinal Loss}
We improve the robustness of the ordinal loss $\mathcal{L}_{ordinal}$ in Equation \ref{equ:logistic_y}, by introducing a non-negative random variable $s_{ij}$ associated with each observed ordinal labels $(i, j) \in \mathcal{J}$, such that
\begin{equation}
    \mathcal{L}_{ordinal}(u_i, u_j, s_{ij}) = - \sum_{(i, j) \in \mathcal{J}}
    \log \psi \left(  \vphantom{\frac{1}{1}} s_{ij} ( u_i - u_j ) \right), 
    \label{equ:L_ordinal}
\end{equation}
where we let $u_n = f(\mathbf{x}_n)$ represent the ordinal information that is captured via a certain function $f$. We call $s_{ij}$ trustworthiness score, such that $s_{ij} = 0$ indicates that the ordinal comparison $t_i > t_j$ is noisy and will be ignored by the model; while $s_{ij} >> 0$ indicates that the comparison is more trustworthy and will impact the model more notably. We analyze how this is achieved in Section \ref{sec:s_analysis}, after we introduce our overall model.

We introduce a prior distribution for $s_{ij}$. In order to facilitate the optimization under VAE framework with reparameterization trick \citep{kingma2013auto}, we introduce an auxiliary random variable $w_{ij}$ and let $s_{ij} = w_{ij}^2$. We let $w_{ij}$ follows a Gaussian distribution
\begin{equation}
     w_{ij} \sim \mathcal{N} (\mu_{w}, \sigma_{w}^2),
\end{equation}
where $\mu_{w}$, $\sigma_{w}^2$ are the parameters for the Gaussian distribution. We dicuss the choice of these hyperparameters in Section \ref{sec:mu_var_w}. Note that since we let $s_{ij}$ be a function of $w_{ij}$, we can get samples of $s_{ij}$ by first sampling $w_{ij}$.

\subsection{ Incorporating VAE Framework}
\label{sec:method_framework}   
Now we present how we incorporate the ordinal loss into the VAE framework. Since an instance $\mathbf{x}_n$ can be reconstructed based on both the information that is relevant and irrelevant to the factor of interest, we divide the latent representation $\mathbf{z}_n$ into two sub-spaces, \ie $\mathbf{z}_n = [ u_n, \mathbf{v}_n ] $. We let $u_n \in \mathbb{R}$ represent the latent variable that corresponds to the factor of interest and explains the pairwise ordinal comparison, as introduced in Equation (\ref{equ:L_ordinal}). We let $\mathbf{v}_n \in \mathbb{R}^{D}$ account for the rest of information that is useful to reconstruct $\mathbf{x}_n$. That is to say, a reconstructed instance $\widetilde{\mathbf{x}}_n$ can be obtained by feeding both $u_n$ and $\mathbf{v}_n$ to the decoder network, which is denoted by $p_{\boldsymbol{\theta}}(\widetilde{\mathbf{x}}| u_n, \mathbf{v}_n)$; while the ordinal loss for each pairwise comparison $t_i > t_j$ can be written as a function of $u_i$ and $u_j$, as defined in Equation (\ref{equ:L_ordinal}).

\begin{figure}[t]
    \centering
    \includegraphics[width=1.00\columnwidth, height=1.00\columnwidth, trim={0cm 7cm 0cm 0cm}, clip]{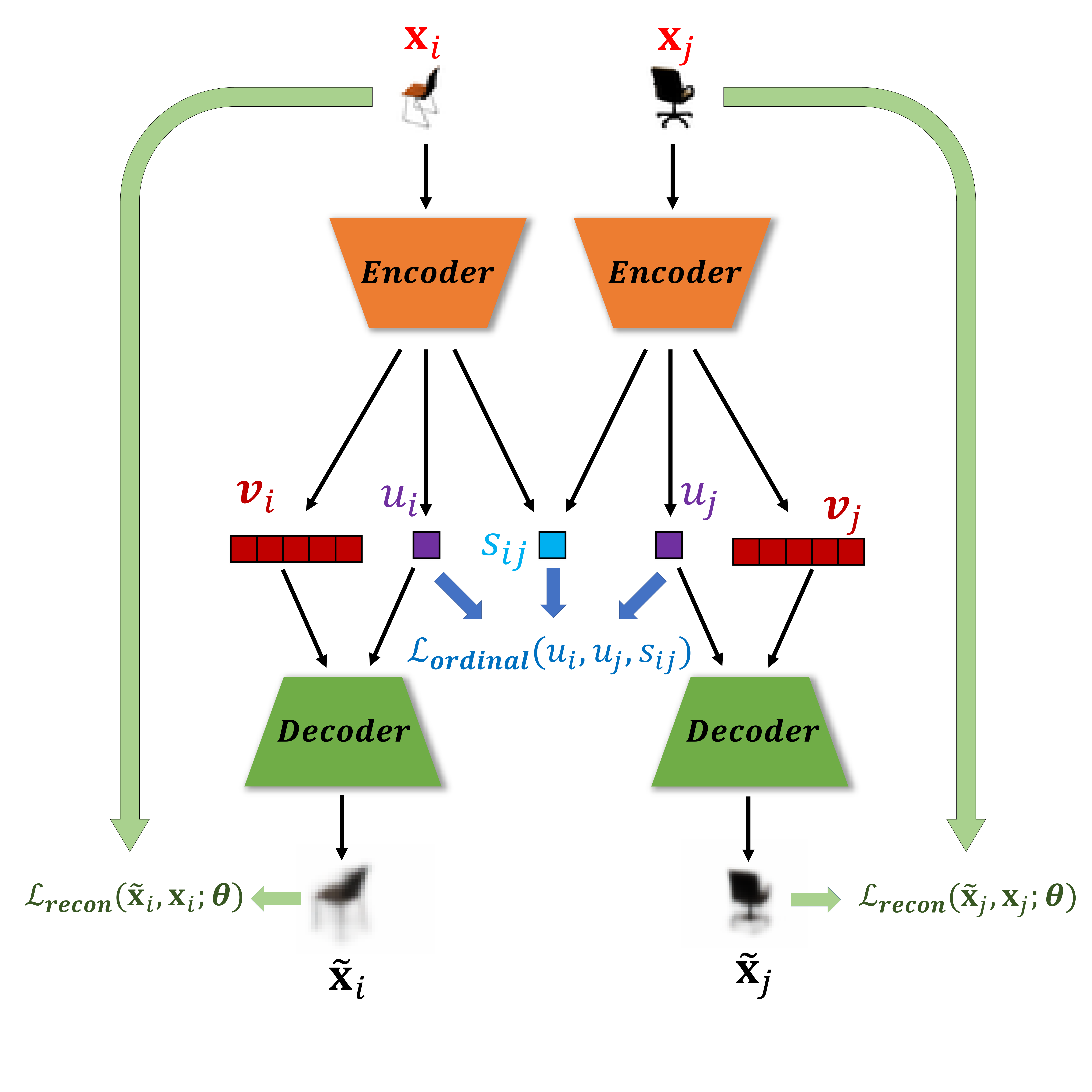}
    \caption{During training, the model makes use of both the observed data $ \{ \mathbf{x} \}_{n=1}^N $ and the pairwise comparison $t_i > t_j$ for each $(i, j) \in \mathcal{J}$. The objective function involves both reconstruction loss $\mathcal{L}_{recon}$ and ordinal loss $\mathcal{L}_{ordinal}$. }
    \label{fig:model}
\end{figure}

\subsection{ Approximating the Posterior Distribution }
In the inference, we need to compute the posterior distribution for all random variables $p( \mathbf{U}, \mathbf{V}, \mathbf{W} | \mathbf{X}, \mathcal{J})$. Since it is intractable to compute the posterior distribution in closed form, we use an encoder network $q_{\boldsymbol{\phi}}(\cdot| \cdot)$ to approximate it, such that it factorizes
\begin{equation}
\begin{aligned}
    p( \mathbf{U}, \mathbf{V}, \mathbf{W} | \mathbf{X}, \mathcal{J}) \approx 
    & \prod_{n=1}^N q_{\boldsymbol{\phi}}( u_n | \mathbf{x}_n )
    q_{\boldsymbol{\phi}}( \mathbf{v}_n | \mathbf{x}_n ) 
    \\
    & \prod_{(i, j) \in \mathcal{J}} q_{\boldsymbol{\phi}}( w_{ij} | \mathbf{x}_i, \mathbf{x}_j )
\end{aligned}
\end{equation}
where $\boldsymbol{\phi}$ is the parameters in the decoder. To facilitate the computation of gradient via reparameterization tricks \citep{kingma2013auto}, we let  $q_{\boldsymbol{\phi}} ( u_n | \mathbf{x}_n )$, $q_{\boldsymbol{\phi}} ( \mathbf{v}_n | \mathbf{x}_n )$ and $q_{\boldsymbol{\phi}} ( \mathbf{w}_{ij} | \mathbf{x}_i, \mathbf{x}_j )$ all follow Gaussian distributions, which is consistent to the standard VAE framework.

\subsection{Overall Model}
The overall objective function can be summarized as follows:
\begin{equation}
    \begin{adjustbox}{width=1.\columnwidth,center}
    {$
    \begin{aligned}
        \mathcal{L} = &
        \min_{ \boldsymbol{\phi}, \boldsymbol{\theta} }
        \mathbb{E}_{\mathbf{x}_n \sim \text{data} }
        [\mathbb{E}_{ q_{\boldsymbol{\phi}} ( \mathbf{z}_n | \mathbf{x}_n ) } 
        [
        \mathcal{L}_{recon} (\widetilde{\mathbf{x}}_n,\mathbf{x}_n ;\boldsymbol{\theta}) ] ]
        \\
        & + \alpha \mathbb{E}_{ (i,j) \sim \mathcal{J} } 
        [\mathbb{E}_{ q_{\boldsymbol{\phi}}(u_i, u_j, w_{ij}| \mathbf{x}_i, \mathbf{x}_j)
        }
        [
         \mathcal{L}_{ordinal}(u_i, u_j, w_{ij}) ] 
        \\
        & + \beta \mathbb{E}_{\mathbf{x}_n \sim \text{data} }
         \left [ \mathcal{D}_{KL} 
         \left( \vphantom{ \frac{1}{1} } 
         q_{\phi}( \mathbf{z} ^{(u)}_n |\mathbf{x}_n) || \mathcal{N}( \boldsymbol{0}, \boldsymbol{I} \right) \right]
        \\
        & + \mathbb{E}_{ (i,j) \sim \mathcal{J} }
        \left[ \mathcal{D}_{KL} 
        \left( \vphantom{ \frac{1}{1} } 
        q_{\phi}( w_{ij} | \mathbf{x}_i, \mathbf{x}_j) || \mathcal{N}( \mu_w,  \sigma_w^2 ) \right) \right].
    \end{aligned}
    $}
    \end{adjustbox}
    \label{equ:objective}
\end{equation}
We visualize the overall model in Figure \ref{fig:model}. The non-negative parameter $\alpha$ in Equation (\ref{equ:objective}) controls the trade off between the ordinal loss and other terms in the objective function. Similar to $\beta$-VAE \citep{higgins2017beta}, we introduce a parameter $\beta$ for the KL divergence term. As mensioned by \citet{mathieu2019disentangling}, introducing $\beta$ helps control the overlap of encodings in the latent space and therefore lead to better generalization of the latent representations. We solve the  optimization problem for $\boldsymbol{\theta}$ and $\boldsymbol{\phi}$ stochastic gradient descent (SGD). 

\subsection{The Impact of Trustworthiness Scores}
\label{sec:s_analysis}
After introducing the overall objective function, now we analyze how the random variables $u_i$, $u_j$ and $w_{ij}$ interact during training. For simplicity, we let the size of minibatch be one. Then in the $(t+1)$ iteration in the training, we first sample $u_i$, $u_j$, $s_{ij}$ for a certain $(i,j)~\in~\mathcal{J}$ from the encoder model $q_{\boldsymbol{\phi}^{(t)}}$ via reparameterization trick, where $\boldsymbol{\phi}^{(t)}$ represents the current parameters for the encoder model. Now we analyze the ordinal loss term in the objective function in Equation~(\ref{equ:objective}), because it defines how $u_i$, $u_j$ and $s_{ij}$ interact. This term can be written as
\begin{equation}
    \label{equ:interaction}
    h(u_i, u_j, s_{ij}) = - \log \psi( s_{ij} (u_i - u_j) )
\end{equation}
Note that $s_{ij}$, $u_i$ and $u_j$ can be written as functions of $\boldsymbol{\phi}$ by making use of the reparameterization trick.

We first analyze how the embedding of $u_i$ and $u_j$ is affected if we regard $s_{ij}$ as a constant. 
We observe in Equation (\ref{equ:interaction}) that if the pairwise comparison $t_i > t_j$ is untrustworthy such that $s_{ij} \rightarrow 0$, since $- \log \psi(0)$ is a constant with respect to $\boldsymbol{\phi}$, we have $\nabla_{\boldsymbol{\phi}}h = 0$. Therefore, $\boldsymbol{\phi}$ will not be changed in the gradient descent update. This indicates that untrustworthy ordinal labels $t_i > t_j$ are ignored by the model. In contrast, if $s_{ij} >>0$, since the logistic function $\psi$ is monotonic increasing, minimizing $h$ is equivalent to maximizing $u_i - u_j$. This implies that if $s_{ij} >> 0$, the embeddings of $u_i$ and $u_j$ are affected more significantly when we update the network $\boldsymbol{\phi}$.  

Now we analyze how $s_{ij}$ will be updated, regarding $u_i$ and $u_j$ are constant, when we minimize Equation~(\ref{equ:interaction}). Because the logistic function $\psi$ is monotonic increasing, we observe that  $s_{ij} \rightarrow \infty$ if $u_i > u_j$, and $s_{ij} \rightarrow 0$ if $u_i < u_j$, when we achieve the minimum $h$.  Since $u_i$ and $u_j$ are sampled from the current encoder model $q_{\boldsymbol{\phi}^{(t)}}$ through the reparameterization trick, we conclude that $s_{ij}$ tend to have large value if the observed ordinal comparison $ t_i > t_j$ is consistent with the current encoder model $q_{\boldsymbol{\phi}^{(t)}}$, ( \ie $u_i > u_j$ ), and vice versa.

\section{ Related work }
As introduced in Section \ref{sec:background_VAE}, several unsupervised disentanglement methods have been previously proposed. However, without explicit supervision, it is difficult to control the correspondence between a learned latent variable and a semantic meaning, and it is not guaranteed that the factor of interest can always be successfully disentangled \citep{locatello2018challenging}. In contrast, our proposed method ROVAE utilizes the ordinal labels as explicit supervision, which encourages the model to disentangle the factor of interest.

There have been attempts to improve disentanglement performance by introducing the supervision of various forms.
\citet{narayanaswamy2017learning} and \citet{kulkarni2015deep} propose semi-supervised VAE methods that learn disentangled representation, by making use of partially observed class labels or real-value targets. \citet{bouchacourt2018multi} introduces supervision via grouping the samples. \citet{feng2018dual, chen2018multiview} and \citet{chen2019weakly} propose to make use of pairwise similarity labels. In contrast, our proposed method utilizes pairwise comparisons as supervision.

% !TEX root = ../main.tex

\section{Experiments}
\label{sec:Experiments}

In this section, we evaluate our method by performing experiments on both benchmark and real-world datasets. In the following, we first introduce the competitive method and the quantitative metrics used in this paper. Then, we introduce the performance of our method when noisy labels are observed in the benchmark datasets. Later, we present the results where we apply our method to analyze a real-world dataset. Finally, we present some results of ablation study.

\subsection{ Competitive Method and Quantitative Metrics }

To demonstrate that ROVAE is more robust to the noise, we compare it with a Vanilla Ordinal VAE (VOVAE). VOVAE shares a similar framework of ROVAE, but fixes $s_{ij} = 1$.
We also compare ROVAE with existing unsupervised disentanglement approaches, including $\beta$-VAE\citep{higgins2017beta}, HCV~\citep{lopez2018information}, factor-VAE~\citep{kim2018disentangling}, $\beta$-TCVAE~\citep{chen2018isolating}.

To measure the performance quantitatively, we compare the methods based on two criteria: Mutual Information Gap (MIG) and regression/classification error. 

MIG is a popular measure for unsupervised disentanglement, first introduced by \cite{chen2018isolating}. Since we introduce supervision in our model, we use a variant of it, defined as
\begin{equation}
    \label{equ:MIG}
    MIG = \frac{1}{ \mathcal{H}(t) } \left( \mathcal{I} (u; t) - \max_{d \in \{1 \ldots D \}} \mathcal{I} (v_d; t) \right),
\end{equation}
where $u$ and $\mathbf{v}$ represent the latent variables that are relevant and irrelevant to the factor of interest, respectively, as introduced in Section~\ref{sec:method_framework} and $t$ denote the ground-truth factor of variations of interest. We let $\mathcal{I}(\cdot;\cdot)$ denote the mutual information between two random variables, and $\mathcal{H}(\cdot)$ be the entropy of a random variable. Both $\mathcal{I}(\cdot;\cdot)$ and $\mathcal{H}(\cdot)$ can be estimated empirically, as explained by \citet{chen2018isolating}. 
MIG in Equation (\ref{equ:MIG}) is defined based on the intuition that we want $u$ to be related to $t$ but each $v_d$ to be independent of $t$. Note that MIG in Equation (\ref{equ:MIG}) can be negative because $\mathcal{I}(u;t)$ might be smaller than $\mathcal{I}(v_d;t)$ with a certain $d \in \{1, \ldots, D\}$, especially when we introduce noise.

In addition to MIG, we also want to investigate how well $u$ predicts $t$. Therefore, we train a $5$ Nearest Neighbour (5-NN) regressor/classifier that predicts $t$ using $u$. We report the r-squared value ($r^2$) for real-value $t$ and Cohen's kappa ($\kappa$) for discrete $t$ in the test dataset. Both $r^2$ and $\kappa$ have a maximum value of $1$ represent a better prediction, and higher value indicates a better performance.

In the experiments, we also compute the quantitative measures for unsupervised disentanglement methods, where $u$ and $\mathbf{v}$ are not defined. For these variables, we pick the dimension of the latent variable that has the maximum mutual information with respect to the ground-truth $t$ in the training set as $u$. The remaining latent variables are regarded as $\mathbf{v}$ in the computation.

\subsection{ Benchmark Datasets }

We evaluate our methods on the following four benchmark datasets:  Yale Faces~\citep{georghiades2001few}, 3D chairs~ \citep{aubry2014seeing} and 3D cars \citep{krause20133d} and UTKFace \citep{zhifei2017cvpr}. The details of these datasets are summarized in Table~\ref{tbl:dataset}. For each dataset, we generate a subset of pairwise ordinal labels based on comparing the ground-truth factor of variations. For each dataset, we randomly sample $5,000$ pairwise labels . Note that for 3D chairs and 3D cars datasets, we only include the instances with azimuth rotations between $0$ to $180$ degrees. We do not include the remaining instances because azimuth is cyclic. The ordinal comparisons between two azimuth values are unclear if we include all instances between $0$ to $360$ degrees.

We want to measure the robustness of the proposed ROVAE against noisy ordinal labels. We consider the noise of two types:
(1) We first generate the pairwise ordinal labels by observing whether the ground-truth factors satisfying $t_i > t_j$. Then we flip the ordinal labels, \ie we let $(j, i) \in \mathcal{J}$,  with probability $0 \le \eta \le 0.5$. Note that we do not consider the case when $\eta>0.5$, because this is equivalent to reversing the ordinal sequences and flipped the ordinal labels with probability $1 - \eta$. This type of noise simulates that the user makes random mistakes in comparing each pair of instances.
(2) We first generate random Gaussian noise $ \epsilon_1,\epsilon_2 \sim \mathcal{N}(0, \sigma^2)$. Then we generate the pairwise ordinal labels by observing whether $\tilde{t}_i + \epsilon_1 > \tilde{t}_j + \epsilon_2$ is true or not, where $\tilde{t}_i$ and $\tilde{t}_j$ are the normalized ground-truth factor of variations with zero mean and unit variance. This type of noise simulates that the user is more likely to make mistakes when $t_i \approx t_j$.

\subsubsection{Quantitative Comparisons}
\begin{figure*}[t]
\begin{minipage}[t]{1.\textwidth}
    \centering
    \begin{subfigure}[t]{.24\textwidth}
        \includegraphics[width=1\textwidth]{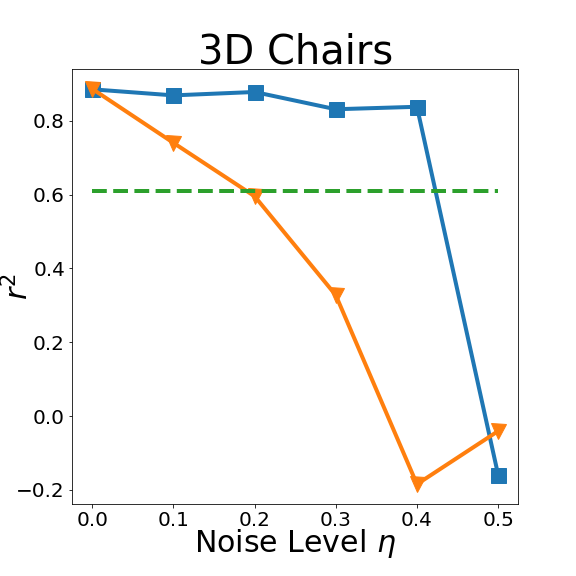}
    \end{subfigure}
    \hfill
    \begin{subfigure}[t]{.24\textwidth}
        \includegraphics[width=1\textwidth]{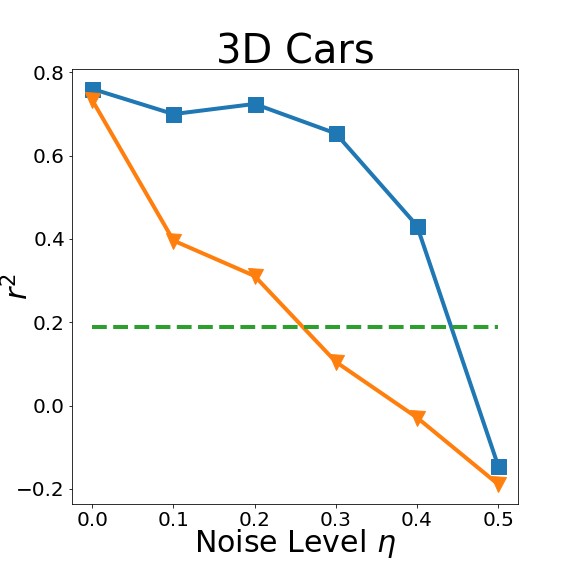}
    \end{subfigure}
    \hfill
    \begin{subfigure}[t]{.24\textwidth}
        \includegraphics[width=1\textwidth]{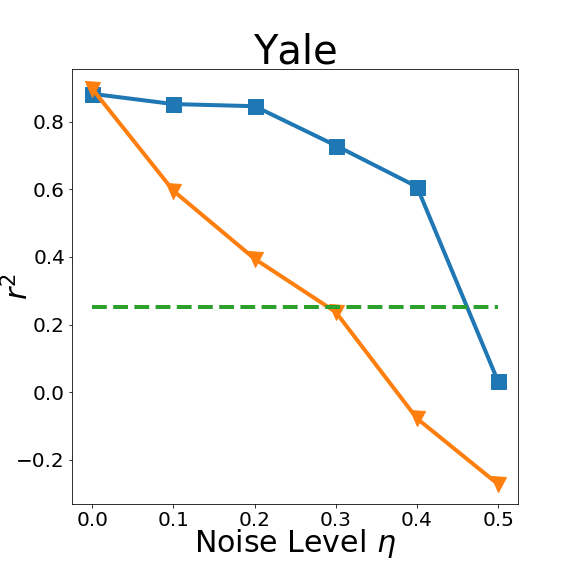}
    \end{subfigure}    
    \hfill
    \begin{subfigure}[t]{.24\textwidth}
        \includegraphics[width=1\textwidth]{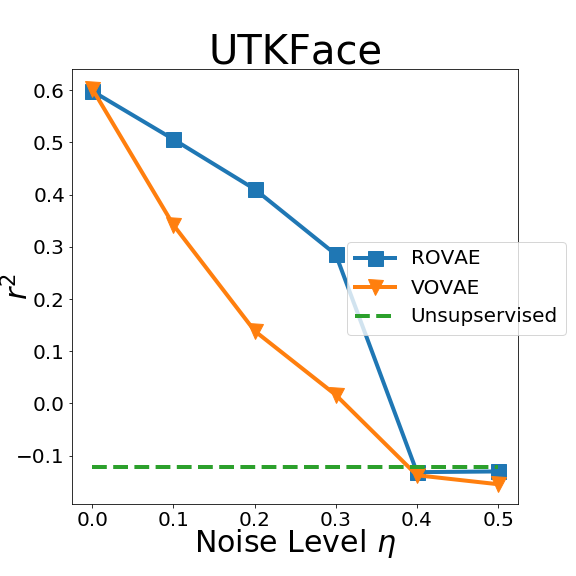}
    \end{subfigure}    
    \\
    \begin{subfigure}[t]{.24\textwidth}
        \includegraphics[width=1\textwidth]{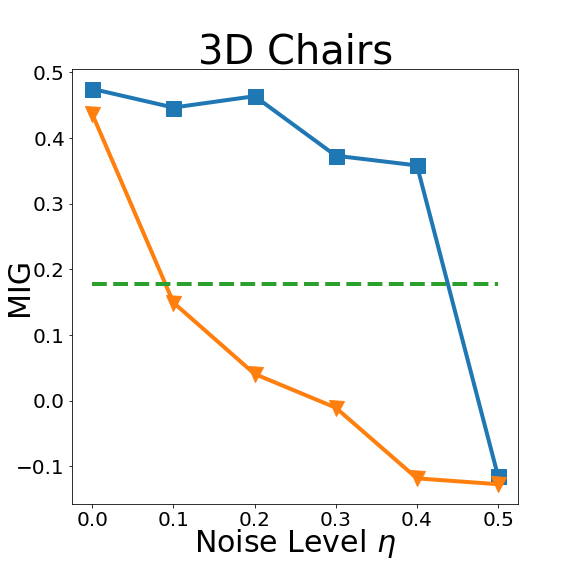}
    \end{subfigure}
    \hfill
    \begin{subfigure}[t]{.24\textwidth}
        \includegraphics[width=1\textwidth]{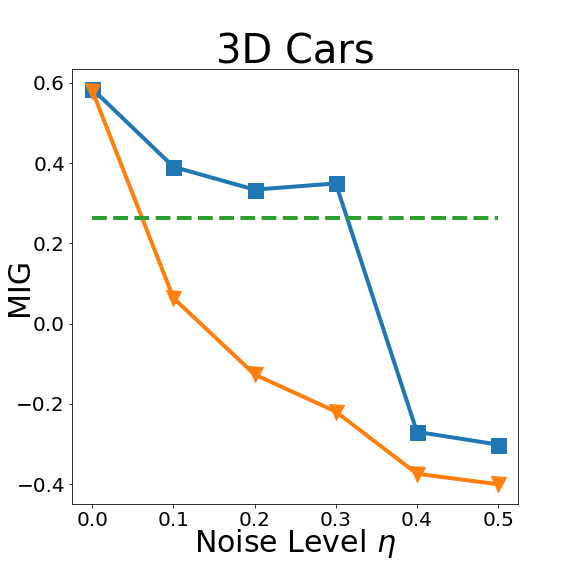}
    \end{subfigure}
    \hfill
    \begin{subfigure}[t]{.24\textwidth}
        \includegraphics[width=1\textwidth]{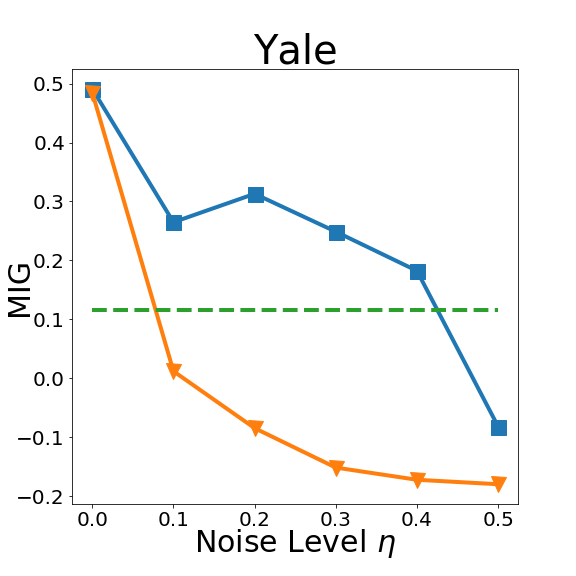}
    \end{subfigure}
    \begin{subfigure}[t]{.24\textwidth}
        \includegraphics[width=1\textwidth]{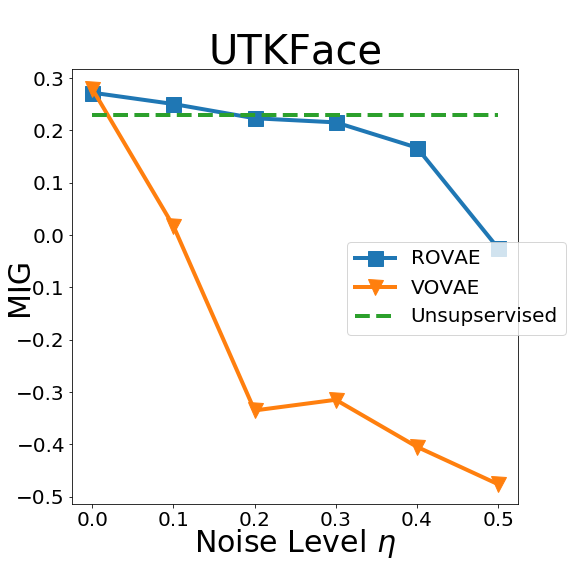}
    \end{subfigure}
    \caption{Plots of $r^2$ and MIG against different noise level $\eta$, where we randomly flip the ground-truth pairwise ordinal labels with probability $\eta$. In the figures, we plot the performance of the unsupervised method that achieves the highest $r^2$ and MIG values among the $4$ unsupervised methods.}
    \label{fig:noise_eta}
\end{minipage}
\begin{minipage}[t]{1.\textwidth}
    \centering
    \begin{subfigure}[t]{.24\textwidth}
        \includegraphics[width=1\textwidth]{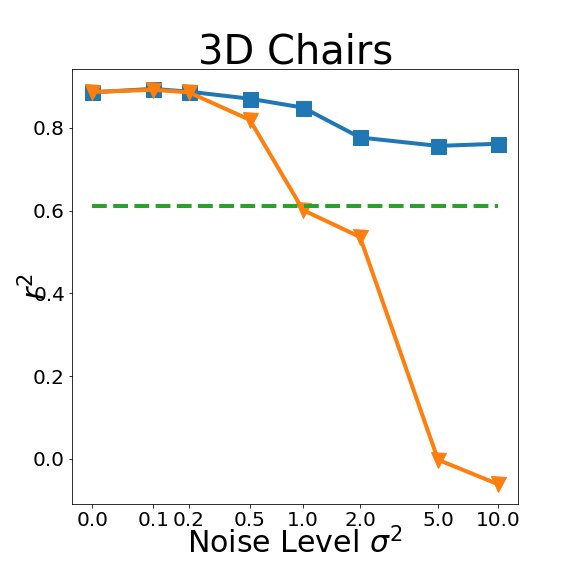}
    \end{subfigure}
    \hfill
    \begin{subfigure}[t]{.24\textwidth}
        \includegraphics[width=1\textwidth]{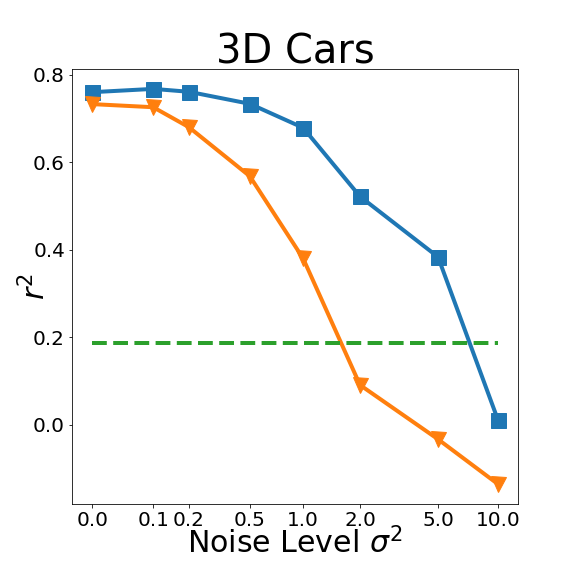}
    \end{subfigure}
    \hfill
    \begin{subfigure}[t]{.24\textwidth}
        \includegraphics[width=1\textwidth]{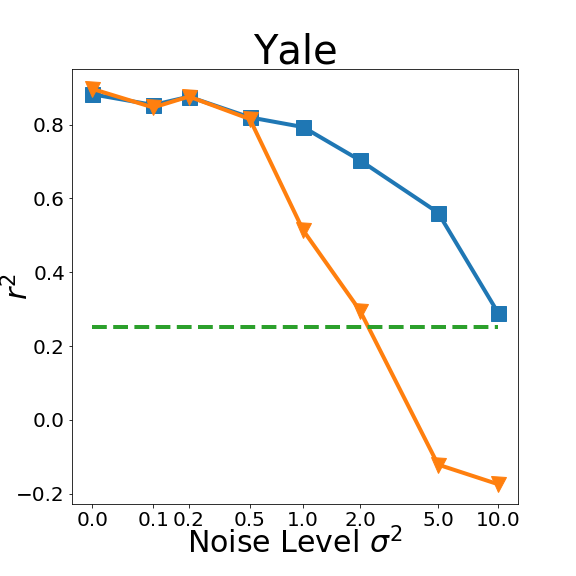}
    \end{subfigure}    
    \hfill
    \begin{subfigure}[t]{.24\textwidth}
        \includegraphics[width=1\textwidth]{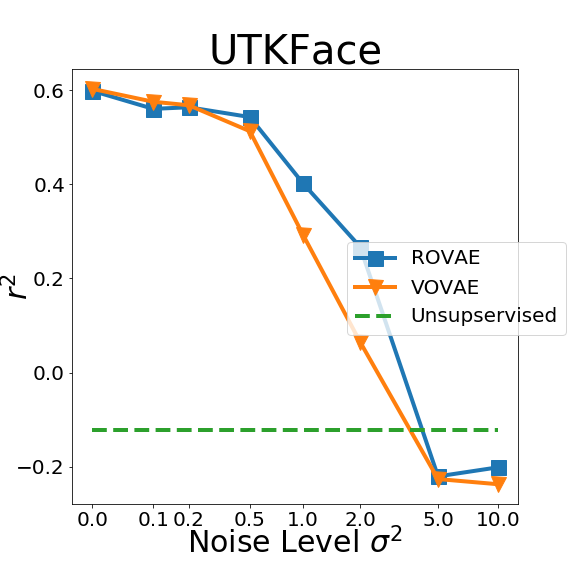}
    \end{subfigure}    
    \\
    \begin{subfigure}[t]{.24\textwidth}
        \includegraphics[width=1\textwidth]{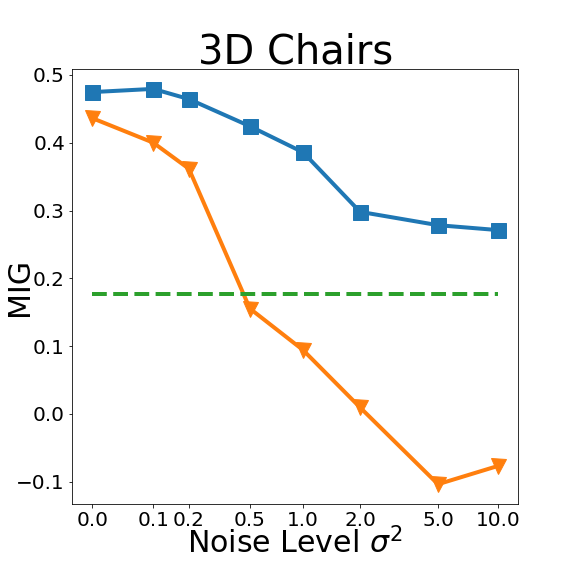}
    \end{subfigure}
    \hfill
    \begin{subfigure}[t]{.24\textwidth}
        \includegraphics[width=1\textwidth]{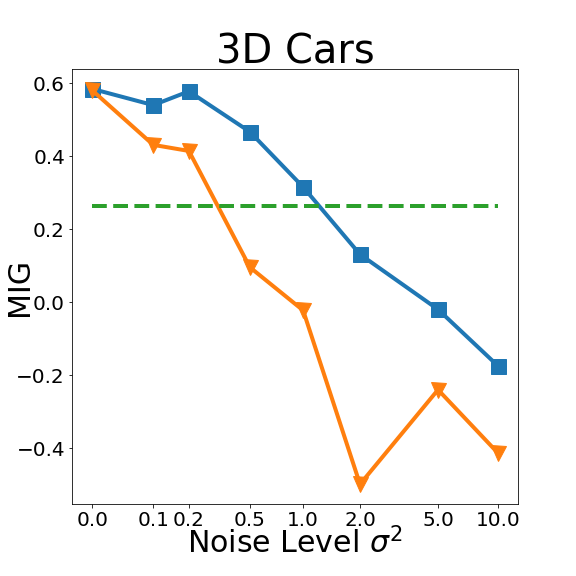}
    \end{subfigure}
    \hfill
    \begin{subfigure}[t]{.24\textwidth}
        \includegraphics[width=1\textwidth]{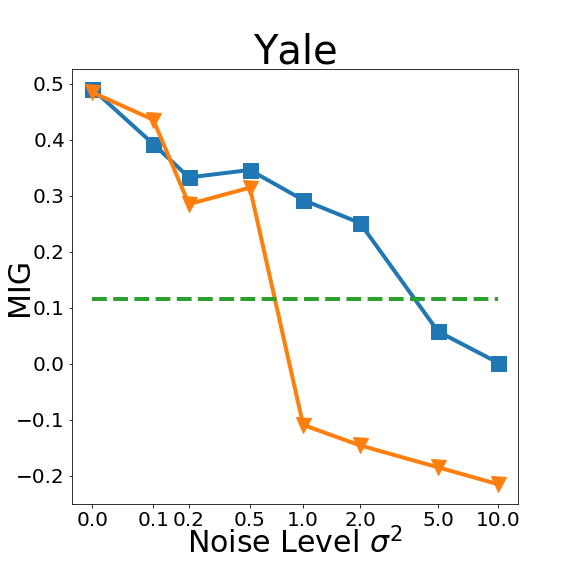}
    \end{subfigure}
    \hfill
    \begin{subfigure}[t]{.24\textwidth}
        \includegraphics[width=1\textwidth]{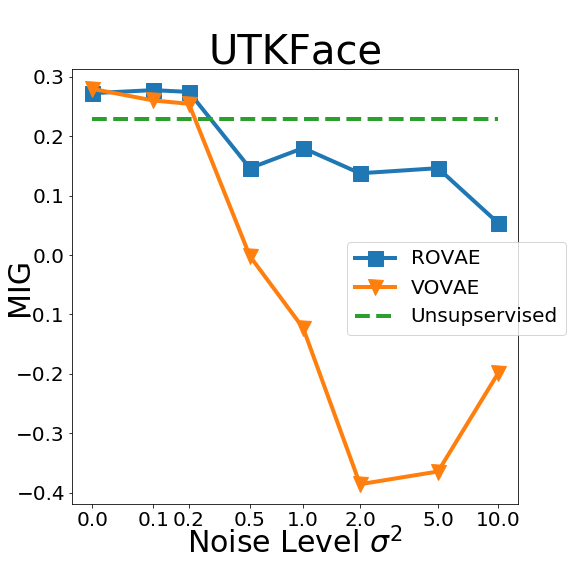}
    \end{subfigure}
    \caption{Plots of $r^2$ and MIG against the noise level $\sigma^2$, where we apply a Gaussian noise with variance $\sigma^2$ to the ground-truth factors before we generate labels based on the pairwise comparisons. In the figures, we plot the performance of the unsupervised method that achieves the highest $r^2$ and MIG values among the $4$ unsupervised methods.}
    \label{fig:noise_sigma}
    \end{minipage}
    \end{figure*}

We plot the MIG and $r^2$ after introducing the two types of noise in Figure \ref{fig:noise_eta} and \ref{fig:noise_sigma}, respectively. We observe in the figures that when the noise level is low, ROVAE does not always outperform VOVAE. However, when the noise level is moderate, ROVAE gives a higher $r^2$ and MIG than VOVAE consistently. When we further increase the noise level, both methods fail. By comparing the performance of ROVAE and VOVAE, we conclude that after introducing the trustworthiness score, ROVAE is able to generate more robust results.

In Figure \ref{fig:noise_eta} and \ref{fig:noise_sigma}, the performance of the unsupervised learning methods are represented using a horizontal line, because these methods do not make use of the labels, and therefore are not affected by the noisy labels. We observe that ROVAE and VOVAE outperform the unsupervised methods when the noise level is low or moderate. The results show that introducing pairwise ordinal labels is helpful in the disentangling tasks, and it is unlikely to disentangle the factor of interest if we do not introduce any supervisions.

\subsubsection{Qualitative Comparisons}
To compare the methods qualitatively when noisy pairwise labels are presented, we generate some plots for the 3D chairs dataset, where $20\%$ of the labels are flipped ($\eta = .2$) as an example. We include the $\beta$VAE in addition to ROVAE and VOVAE, because it is the best unsupervised method for this dataset.

\begin{figure}[t!]
    \hfill
    \begin{subfigure}[t]{.48\columnwidth}
        \includegraphics[width=1\textwidth, 
                         height=1.\textwidth, trim={0cm 1cm 1cm 2cm}, clip]
        {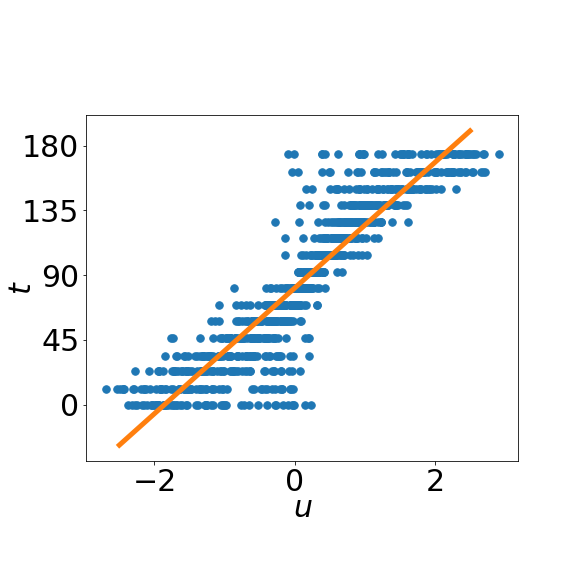}
        \caption{The plot of $t$ versus $u$ for ROVAE.}
        \label{fig:embedding_ROVAE}
    \end{subfigure}
    \hfill
    \begin{subfigure}[t]{.48\columnwidth}
        \includegraphics[width=1\textwidth, 
                         height=1.\textwidth,]{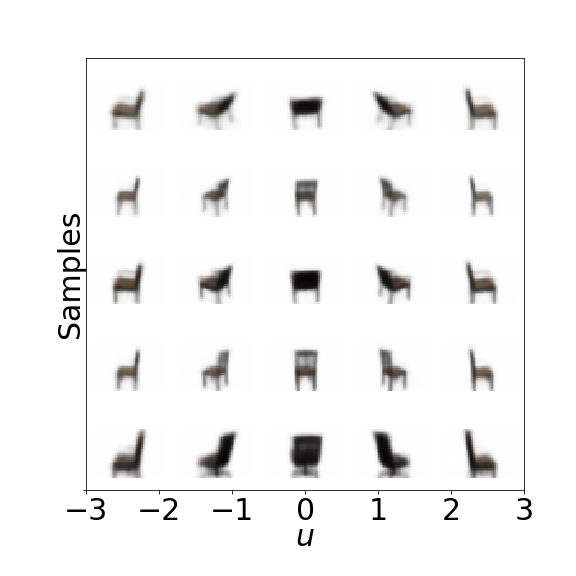}
        \caption{Generated images of ROVAE by manipulating $u$.}
        \label{fig:generated_ROVAE}
    \end{subfigure}    
    \hfill
    \\
    \hfill
    \begin{subfigure}[t]{.48\columnwidth}
        \includegraphics[width=1\textwidth, 
                         height=1.\textwidth,  trim={0cm 2cm 1cm 2cm}, clip]
        {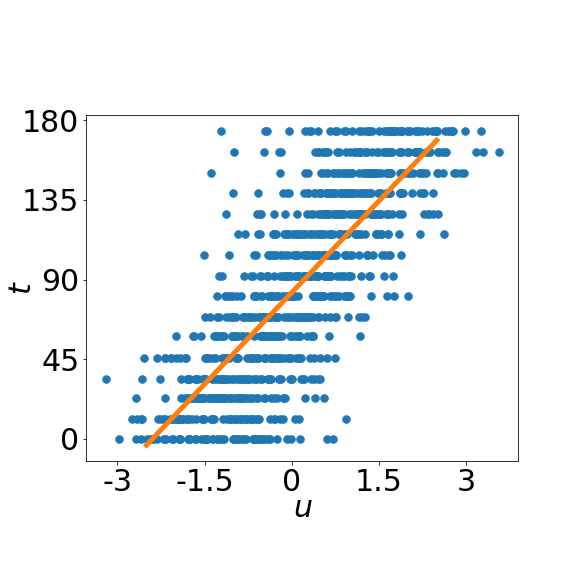}
        \caption{The plot of $t$ versus $u$ for VOVAE.}
        \label{fig:embedding_VOVAE}
    \end{subfigure}
    \hfill
    \begin{subfigure}[t]{.48\columnwidth}
        \includegraphics[width=1\textwidth, 
                         height=1.\textwidth, ]{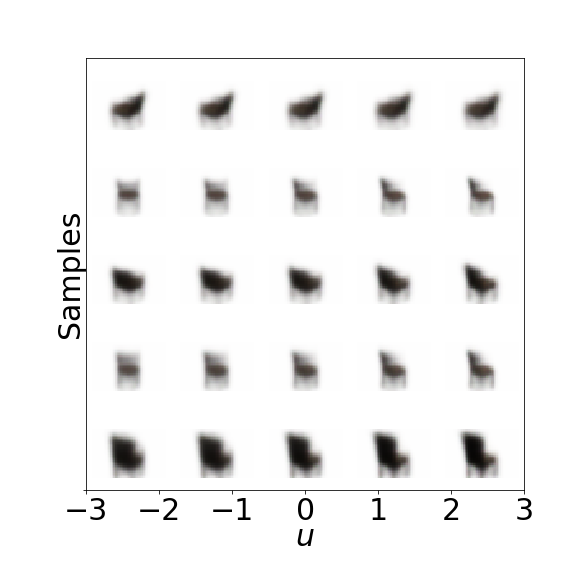}
        \caption{Generated images of VOVAE by manipulating $u$.}
        \label{fig:generated_VOVAE}
    \end{subfigure}        
    \hfill
    \\
    \hfill
    \begin{subfigure}[t]{.48\columnwidth}
        \includegraphics[width=1\textwidth, 
                         height=1.\textwidth,  trim={0cm 2cm 1cm 2cm}, clip]
        {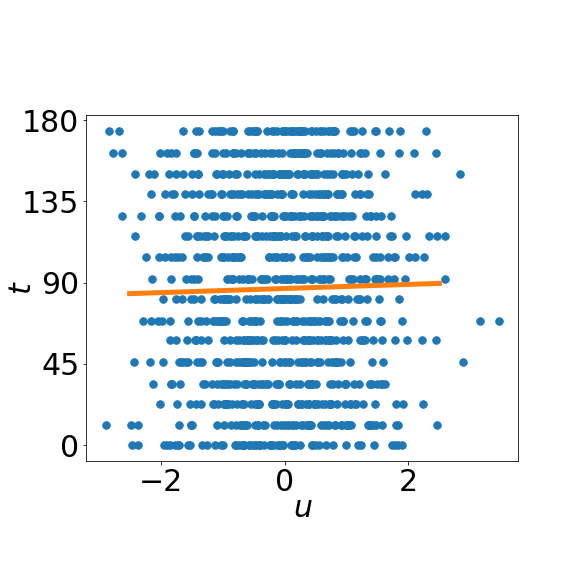}
        \caption{The plot of $t$ versus $u$ for $\beta$VAE.}
        \label{fig:embedding_betaVAE}
    \end{subfigure}
    \hfill
    \begin{subfigure}[t]{.48\columnwidth}
        \includegraphics[width=1\textwidth, 
                         height=1.\textwidth, ]{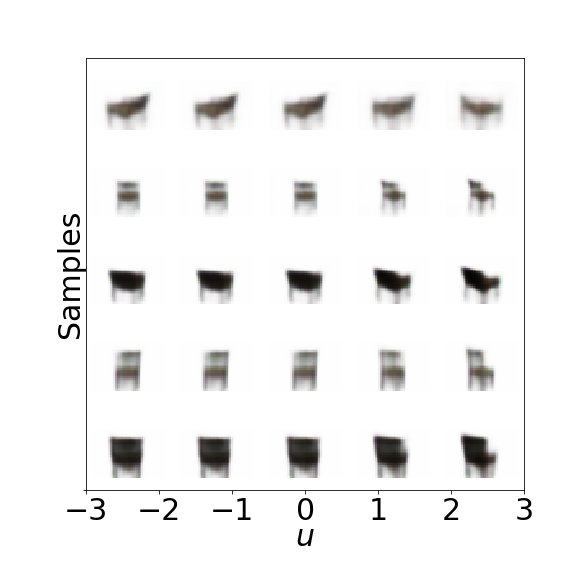}
        \caption{Generated images of $\beta$VAE by manipulating $u$.}
        \label{fig:generated_betaVAE}
    \end{subfigure}   
    \hfill
    \caption{Embedding and reconstruction results for the 3D chairs dataset, where $20\%$ of the labels are flipped ($\eta = .2$).}
    \vspace{-3mm}
\end{figure}

To visualize whether or not the methods embed the instances based on the factor of interest, we plot $t$ versus $u$ for the held-out data in Figure \ref{fig:embedding_ROVAE}, \ref{fig:embedding_VOVAE} and \ref{fig:embedding_betaVAE}. In the plots, we also include the optimal linear model that predicts $t$ based on $u$. As shown in Figure \ref{fig:embedding_ROVAE}, the latent representation $u$ generated by ROVAE can better predict $t$ when $45 < t < 135$, although larger error is observed when predicting $t$ with other values. In contrast, VOVAE is more affected by the noise introduced, as shown in Figure \ref{fig:embedding_VOVAE}. We observe the prediction less accurate. Although $\beta$VAE is not affected by the label noise, it does not give good embedding, as shown in Figure \ref{fig:embedding_betaVAE}. Without introducing supervision, it is not guaranteed that unsupervised methods can recover the ground-truth factor of variations. By comparing Figure \ref{fig:embedding_ROVAE}, \ref{fig:embedding_VOVAE} and \ref{fig:embedding_betaVAE}, we conclude that the embedding results of ROVAE are more consistent with the ground-truth factor of interest in this task.

To illustrate whether or not the models disentangle the factor of interest and represent it with the latent representation $u$, we encode the instances into $[u, \mathbf{v}]$, keep $\mathbf{v}$ constant but manipulate $u$ and then generate images by decoding the results. We include some results in Figure \ref{fig:generated_ROVAE}, \ref{fig:generated_VOVAE} and \ref{fig:generated_betaVAE}. In Figure \ref{fig:generated_ROVAE}, we observe that the chairs generated by ROVAE rotate when we change the values of $u$. However, due to the introduced noise, the chairs generated by VOVAE do not rotate as much, as shown in \ref{fig:generated_VOVAE}. We can also observe the rotation of generated chairs in Figure \ref{fig:generated_betaVAE}, indicating $\beta$VAE is able to disentangle the factor of interest to a certain degree. However, the results are not as good as those generated by ROVAE. By comparing Figure \ref{fig:generated_ROVAE}, \ref{fig:generated_VOVAE} and \ref{fig:generated_betaVAE}, we conclude that ROVAE is a better approach to disentangle the factor of interest when noisy ordinal comparisons are presented.

\begin{table*}[t]
\caption{The Benchmark Dataset}
\label{tbl:dataset}

\centering
% \begin{adjustbox}{width=1.\textwidth, center}  
  \begin{tabular}{lcccl}
    \toprule
    Dataset   & Training instances & Held-out instances & Image size & The ground-truth factor \\
    \midrule
    Yale Faces & $1,903$ & $513$& $64 \times 64 \times 1$ & Lighting Angles
    \\
    3D chairs & $69,131$ & $17,237$ & $64 \times 64 \times 3$ & Azimuth Rotations ( $0$\textemdash$180 ^{\circ} $ )
    \\
    3D cars & $14,016$ & $3,552$ & $128 \times 128 \times 3$ & Azimuth Rotations  ( $0$\textemdash$180 ^{\circ} $ )
    \\
    UTKFace & $37,932$ & $9,484$ & $200 \times 200 \times 3$ & Ages
    \\
    \bottomrule
  \end{tabular}
% \end{adjustbox}  
\end{table*}

\begin{figure}[t]
    \begin{subfigure}[t]{.48\columnwidth}
        \includegraphics[width=1\textwidth, height=1.15\textwidth, 
                         trim={1cm 2cm 0cm 0cm}, clip ]{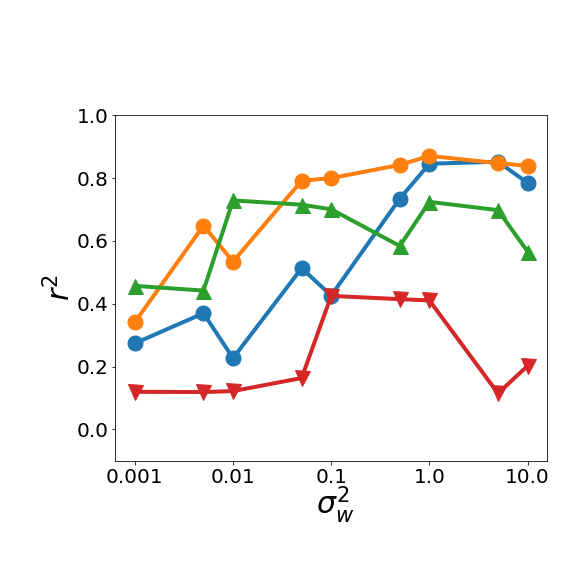}
        \caption{ We fix $\mu_w = 10$, variate $\sigma^2_w$ and plot $r^2$ for different datasets.}
        \label{fig:par_var}
    \end{subfigure}
    \hfill
    \begin{subfigure}[t]{.48\columnwidth}
        \includegraphics[width=1\textwidth, height=1.15\textwidth,
                         trim={1cm 2cm 0cm 0cm}, clip ]{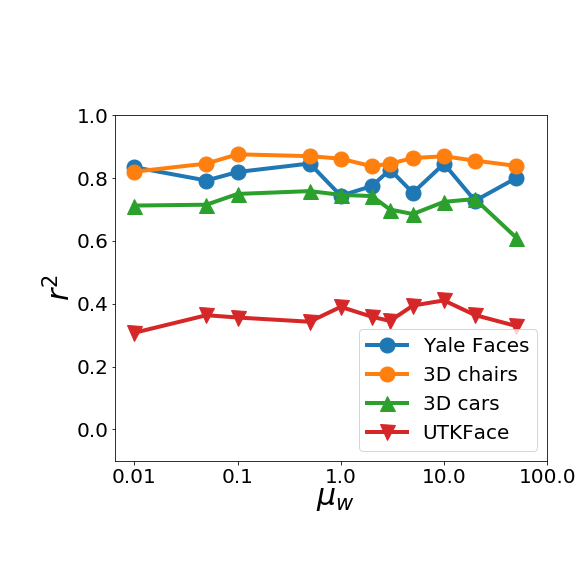}
        \caption{  We fix $\sigma^2_w$ = 1, variate $\mu_w$ and plot $r^2$ for different datasets. }
        \label{fig:par_mu}
    \end{subfigure}    
    \caption{ We plot how the parameters $(\mu_w, \sigma^2_w)$ affects the performance of ROVAE when noisy pairwise ordinal labels are observed. In this experiment, we randomly flipped $20\%$ of the pairwise ordinal pairs (\ie $\eta = .2$) in each dataset. }
    \label{fig:hyperparameter}
\end{figure}

\begin{figure}[t]
    \centering
    \includegraphics[width=.8\columnwidth, height = .8 \columnwidth, trim={.5cm 2cm .5cm 1cm}, clip ]{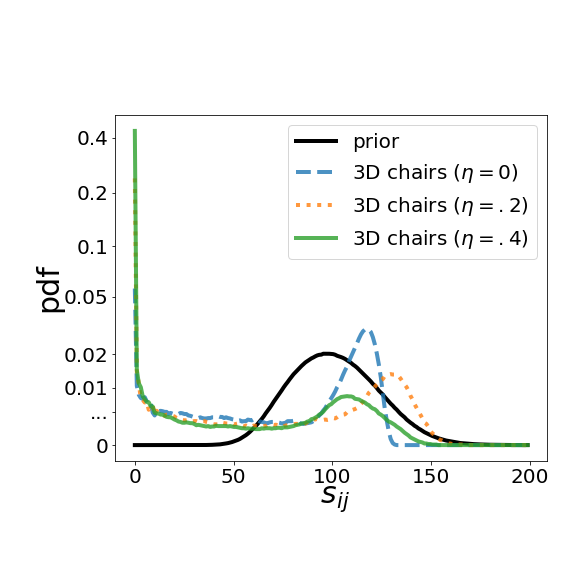}
    \caption{The aggregated posterior distribution $q(s_{ij})$, where the $y$ axis is plotted in log scale. The posterior distribution is a mixture of two components. One component is an impulse located close to $0$ and the other has a mode that is greater than $100$. }
    \label{fig:posterior}
\end{figure}

\subsection{ The Clinical Dataset }

\begin{table*}[t]
    \centering
    \caption{Results on the COPD Dataset}
    \label{tbl:COPD}
    \begin{tabular}{l l c c c c c c c}
    \toprule
     & & ROVAE & VOVAE & $\beta$VAE & HCV & Factor VAE & $\beta$TCVAE & GOLD criteria
    \\
    \midrule
    $\kappa$ & GOLD criteria
    & $.663$ & $.542$ & $.149$ & $.154$ & $.241$ & $.165$ & $\mathbf{1.00}$
    \\
    \midrule
    \multirow{2}{*}{$r^2$}  & \% Emphysema
    & $\mathbf{.491}$ & $.322$ & $-.215$ & $-.070$ & $-.226$ & $.047$ & $.244$
    \\
    & \% Gas Trapped
    & $\mathbf{.558}$ & $.401$ & $-.159$ & $-.089$ & $-.249$ & $.123$ & $.398$
    \\
    \bottomrule
    \end{tabular}
    
    \label{tab:my_label}
\end{table*}

A real-world application of ROVAE is to analyze the severity of Chronic Obstructive Pulmonary Disease (COPD), which is a devastating disease related to cigarette smoking. In clinical practice, the GOLD criteria \citep{singh2019global} is widely used to classify COPD patients into $5$ ordinal categories from healthy to very severe COPD, based on the severity of airflow limitation. Now that there are chest Computer Tomography (CT) images available, we want to investigate how CT images are related to disease severity. We make use the GOLD criteria to generate pairwise ordinal labels to guide the analysis for CT images during training. As demonstrated by \citet{matsuoka2010quantitative}, the CT appearance of the lungs can look similar in terms of severity, even in patients with very different levels of airflow limitations. This implies that the generated ordinal labels are noisy.

In addition to the GOLD criteria, we also train 5~-~NN models using the learned latent variable $u$ to predict following two measures for disease severity including (1)~\texttt{Emphesyma\%} that measures the percentage of destructed lung tissue and (2)~\texttt{GasTrap\%} that amounts gas trapped in the lung. We report $\kappa$ and $r^2$ measured with the held-out data in Table \ref{tbl:COPD}. We observe that ROVAE outperforms VOVAE and all unsupervised methods in terms of predicting these measurements. In the table, we also include the prediction performance using the GOLD criteria for comparison. We observe that the severity variable learned by ROVAE gives a better prediction for Emphesyma\% and GasTrap\% compared with the GOLD criteria. This suggests that ROVAE does not simply learn a latent representation that predicts the GOLD criteria as accurately as possible. By ignoring a subset of noisy ordinal labels, ROVAE discovers a better measurement for disease severity under other criteria based on the CT image.

\subsection{ Prior and Posterior Distribution for the Trustworthiness Score}
\label{sec:mu_var_w}

ROVAE introduces random variables $s_{ij}$ in the model. In this subsection, we first discuss how we choose parameters for its prior distribution. We observe that if we do not introduce noise in the ordinal pairs, the performance of the ROVAE is not sensitive to the choice of hyper-parameters $\mu_w$ and $\sigma_w^2$. Therefore, we focus on discussing the case when moderate noise is presented (\ie $\eta = .2$). We variate one of the hyper-parameter when fixed the other and plot the $r^2$ versus one of the hyper-parameters in Figure \ref{fig:par_var} and \ref{fig:par_mu}. We observe that the performance of ROVAE depends less on the choice of $\mu_w$ but is more sensitive to $\sigma_w^2$. When $\sigma_w^2$ is too small, ROVAE degrades into VOVAE, where $s_{ij} \approx \mu_{w}$ for all $(i, j) \in \mathcal{J}$. The model is misled by the noisy ordinal labels, which hurts its performance. In contrast, too large $\sigma_w^2$ allows ROVAE to ignores too many ordinal labels, such that $s_{ij} \approx 0 $ for all $(i, j) \in \mathcal{J}$. This also deteriorates the performance of the model. In all other experiments, we let $\mu_w = 10$ and $\sigma_w^2 = 1$.

Then we discuss how its aggregated posterior distribution $q(s_{ij}) = \frac{1}{|\mathcal{J}|}\sum_{(i, j) \in \mathcal{J}} q(s_{ij} | \mathbf{x}_i, \mathbf{x}_j )$ is affected by the introduced noise. We take the 3D chairs dataset as an example, and estimate $q(s_{ij})$ via sampling. We plot $q(s_{ij})$ with different noise levels in Figure \ref{fig:posterior}. We observe that the posterior $q(s_{ij})$ is a mixture of two components. One component is an impulse located close to $0$ and the other has a mode that is greater than $100$. As introduce in Section \ref{sec:s_analysis}, the ordinal labels $t_i > t_j$ is ignored when $s_{ij} = 0$. We observe that $P(s_{ij} < 1)$ is $5.6\%$, $23.7\%$ and $42.1\%$ when $\eta = 0$, $\eta = .2$ and $\eta = .4$, respectively. When the ordinal supervision is more noisy, more ordinal labels are automatically ignored by ROVAE.
\section{ Conclusion}
In this paper, we focus on a problem setting where the user provides noisy pairwise ordinal comparisons between instances based on a factor to be disentangled. 
We propose a new method we call Robust Ordinal VAE (ROVAE) to learn the disentanglement of latent variables of interest by making use of the observed pairwise ordinal comparison. We introduce a non-negative random variable in the model, such that ROVAE can automatically determine whether each pairwise ordinal label is trustworthy, and the noisy labels will be ignored by the model. Experimental results demonstrate that ROVAE is more robust to noisy labels in both benchmark datasets and a real-world application.

\clearpage
\bibliography{iclr2020_conference}
\bibliographystyle{iclr2020_conference}

\end{document}